\documentclass[10pt,twocolumn,letterpaper]{article}

\usepackage{iccv}              

\usepackage{url}            
\usepackage{booktabs}       
\usepackage{amsfonts}       
\usepackage{nicefrac}       
\usepackage{microtype}      
\usepackage{xcolor}         
\usepackage{comment}
\usepackage{xspace}
\usepackage{amsmath}
\usepackage{amssymb}
\usepackage{multicol}
\usepackage{multirow}
\usepackage{algorithm}
\usepackage{algpseudocode}
\usepackage{comment}
\usepackage{colortbl}
\usepackage{array}
\usepackage[inkscape]{svg}
\usepackage{placeins}

\usepackage{colortbl}
\usepackage{xcolor}
\usepackage{makecell}

\definecolor{first}{RGB}{107,174,214}  
\definecolor{second}{RGB}{158,202,225} 
\definecolor{third}{RGB}{198,219,239}  

\definecolor{iccvblue}{rgb}{0.21,0.49,0.74}
\usepackage[pagebackref,breaklinks,colorlinks,allcolors=iccvblue]{hyperref}
\usepackage{orcidlink}

\def\paperID{16} 
\def\confName{ICCV}
\def\confYear{2025}

\newcommand{\methname}{CoZAD\xspace}

\title{A Contrastive Learning-Guided Confident Meta-learning for Zero Shot Anomaly Detection}

\author{Muhammad Aqeel$^1$\orcidlink{0009-0000-5095-605X} \and Danijel Sko\v{c}aj$^2$\orcidlink{0000-0002-5290-4736} \and Marco Cristani$^{1,3}$\orcidlink{0000-0002-0523-6042} \and Francesco Setti$^{1,3}$\orcidlink{0000-0002-0015-5534} \\
$^1$Dept. of Engineering for Innovation Medicine, University of Verona\\
Strada le Grazie 15, Verona, Italy\\
$^2$Faculty of Computer and Information Science, University of Ljubljana\\
Kongresni trg 12, Ljubljana, Slovenia\\
$^3$Qualyco S.r.l., Strada le Grazie 15, Verona, Italy\\
{\tt\small Contact author: muhammad.aqeel@univr.it}
}

\begin{document}
\maketitle
\begin{abstract}
Industrial and medical anomaly detection faces critical challenges from data scarcity and prohibitive annotation costs, particularly in evolving manufacturing and healthcare settings. To address this, we propose \methname, a novel zero-shot anomaly detection framework that integrates soft confident learning with meta-learning and contrastive feature representation. Unlike traditional confident learning that discards uncertain samples, our method assigns confidence-based weights to all training data, preserving boundary information while emphasizing prototypical normal patterns. The framework quantifies data uncertainty through IQR-based thresholding and model uncertainty via covariance-based regularization within a Model-Agnostic Meta-Learning. Contrastive learning creates discriminative feature spaces where normal patterns form compact clusters, enabling rapid domain adaptation. Comprehensive evaluation across 10 datasets spanning industrial and medical domains demonstrates state-of-the-art performance, outperforming existing methods on 6 out of 7 industrial benchmarks with notable improvements on texture-rich datasets (99.2\% I-AUROC on DTD-Synthetic, 97.2\% on BTAD) and pixel-level localization (96.3\% P-AUROC on MVTec-AD). The framework eliminates dependence on vision-language alignments or model ensembles, making it valuable for resource-constrained environments requiring rapid deployment.
\end{abstract}

\noindent \textbf{Keywords:} Zero-Shot Anomaly Detection, Contrastive Learning, Meta Learning, Soft Confident Learning    
\section{Introduction}
\label{sec:intro}

Industrial anomaly detection faces critical challenges from data scarcity and the prohibitive costs of expert annotation, particularly as manufacturers rapidly introduce new product variants~\cite{bergmann2020uninformed, huang2022registration, you2022unified, aqeel2025meta}. Traditional approaches requiring extensive retraining for each product create unacceptable deployment delays and resource burdens, especially for small and medium enterprises. This problem is particularly acute in scenarios where normal patterns exhibit substantial variation, making it difficult to distinguish between genuine anomalies and acceptable variations in product appearance~\cite{bergmann2019mvtec}.

Similar challenges exist in medical imaging applications, where anomaly detection plays a crucial role in areas such as disease diagnosis, pathology identification, and medical screening~\cite{bao2024bmad, huang2024adapting, zhao2021anomaly}. In healthcare settings, the scarcity of abnormal samples, strict patient privacy regulations, and the high cost of expert annotations from medical specialists further complicate the development of robust anomaly detection systems. Consequently, there is a growing need for methods that can effectively identify anomalies without requiring extensive labeled datasets or domain-specific retraining.

Recent advancements in unsupervised anomaly detection methods~\cite{aqeel2025CoMet, chen2024unified, he2024learning, yao2024glad} have significantly improved detection performance across both industrial and medical domains but still rely on the availability of substantial nominal samples. Moreover, these approaches often struggle with two fundamental limitations: First, they lack the ability to distinguish between genuine anomalies and rare but normal variations, and Second, they provide limited interpretability regarding why a particular sample is classified as anomalous.

Zero-shot anomaly detection (ZSAD) has emerged as a promising direction to address these challenges by enabling cross-domain knowledge transfer without requiring anomalous samples or extensive retraining~\cite{cao2024adaclip, jeong2023winclip, zhu2024llms}. However, ZSAD approaches often suffer from poor generalization when transferring across domains with different visual characteristics and struggle with robustness to variations in normal patterns. Specifically, these methods fail to adequately handle uncertain samples that lie on decision boundaries—either discarding potentially valuable information or allowing noisy samples to degrade model performance. This limitation is particularly problematic in industrial settings where normal patterns can vary significantly due to lighting conditions, manufacturing tolerances, or material variations.

To address these limitations, we propose \emph{\methname}, a novel framework that combines soft confident learning with meta-learning and contrastive feature representation to enhance anomaly detection performance without requiring anomalous samples. Unlike traditional confident learning approaches that completely remove suspicious samples from the training process, \emph{\methname}'s soft confident learning approach assigns confidence-based weights to all data points. This approach preserves valuable boundary information that would otherwise be lost through hard pruning methods, while still allowing the model to focus primarily on the most prototypical nominal samples. By quantifying both model uncertainty through covariance-based metrics and data uncertainty through statistical thresholding based on interquartile ranges, \emph{\methname} adaptively weights each sample's contribution during training according to its estimated confidence score. This design enables the framework to not only detect anomalies but also provide insights into why specific regions or patterns are flagged as anomalous, thereby enhancing interpretability and trustworthiness for critical applications.

Our main contributions are summarized as follows:
\begin{itemize}
   \item A novel soft confident learning framework for anomaly detection that, instead of discarding potential anomalies, assigns confidence-based weights to all samples during training. This approach quantifies both data and model uncertainty to create a more nuanced learning process that preserves valuable boundary information while focusing on prototypical normal patterns.
   
   \item An innovative integration of contrastive learning with confident meta-learning that creates a more discriminative feature space where normal patterns form compact clusters, making anomalies more easily identifiable as outliers.
   
   \item A meta-learning approach that enables rapid adaptation to new domains while maintaining robustness to potential anomalies, effectively balancing task-specific learning with broader generalizability.
   
   \item Comprehensive experimental validation demonstrating that \emph{\methname} outperforms state-of-the-art methods across multiple benchmarks, including industrial inspection and medical imaging applications, while providing improved interpretability and robustness to domain shifts.
\end{itemize}

\section{Related Work}
\label{sec:relatedwork}

Recent advances in zero-shot anomaly detection (ZSAD) have significantly transformed industrial inspection methodologies by enabling anomaly detection without requiring target domain training data. The earliest work, WinCLIP~\cite{jeong2023winclip}, pioneered this transformation by leveraging vision-language models with compositional prompt ensembles and multi-level feature extraction to aggregate classification results obtained from aligning text with sub-images at multiple scales. Further innovations emerged with APRIL-GAN~\cite{chen2023april}, which extends CLIP with specialized linear layers to better align image features with text embeddings in the joint space. Similarly, CLIP-AD~\cite{chen2024clip} uses trainable adapter layers to map fine-grained patch features into a joint embedding space, further employing feature surgery to mitigate opposite predictions.

Additionally, several existing studies have integrated multiple foundation models, such as SAM~\cite{kirillov2023segment} and GroundingDINO~\cite{liu2024grounding}, to collectively enhance ZSAD performance, as demonstrated in SAA/SAA+~\cite{cao2023segment} and ClipSAM~\cite{li2025clipsam}. The field continues to evolve rapidly with contributions from multimodal approaches. Anomaly-OneVision~\cite{xu2025towards} has recently emerged as a specialist visual assistant for ZSAD that incorporates a Look-Twice Feature Matching mechanism to emphasize suspicious anomalous visual tokens, while other researchers have enhanced anomaly localization by integrating zero-shot segmentation models with foreground instance segmentation, enabling more precise detection of subtle defects~\cite{baugh2023zero}.

An increasing number of prompt optimization-based methods have also been proposed. AnomalyCLIP~\cite{zhou2023anomalyclip} and Filo~\cite{gu2024filo} insert learnable vectors into the input text or CLIP encoder layers to avoid extensive engineering on hand-crafted prompt design. AdaCLIP~\cite{cao2024adaclip} and VCP-CLIP~\cite{qu2024vcp} further utilize textual and visual hybrid prompts to enhance CLIP's anomaly perception capability, while Bayesian Prompt Flow Learning~\cite{qu2025bayesian} addresses limitations of manual prompt engineering by modeling the prompt space as a learnable probability distribution, achieving state-of-the-art results across industrial and medical datasets. These advancements collectively demonstrate that ZSAD is becoming increasingly practical for real-world industrial inspection scenarios where labeled anomaly data is scarce or unavailable. Different from these methods, our \methname employs a confidence-guided discriminative approach that enables more effective zero-shot transfer across domains without requiring elaborate vision-language alignments or multi-model ensembles.

\section{\methname FrameWork}
\label{sec:method}

\begin{figure*}[t]
    \centering
    \includegraphics[width=\linewidth]{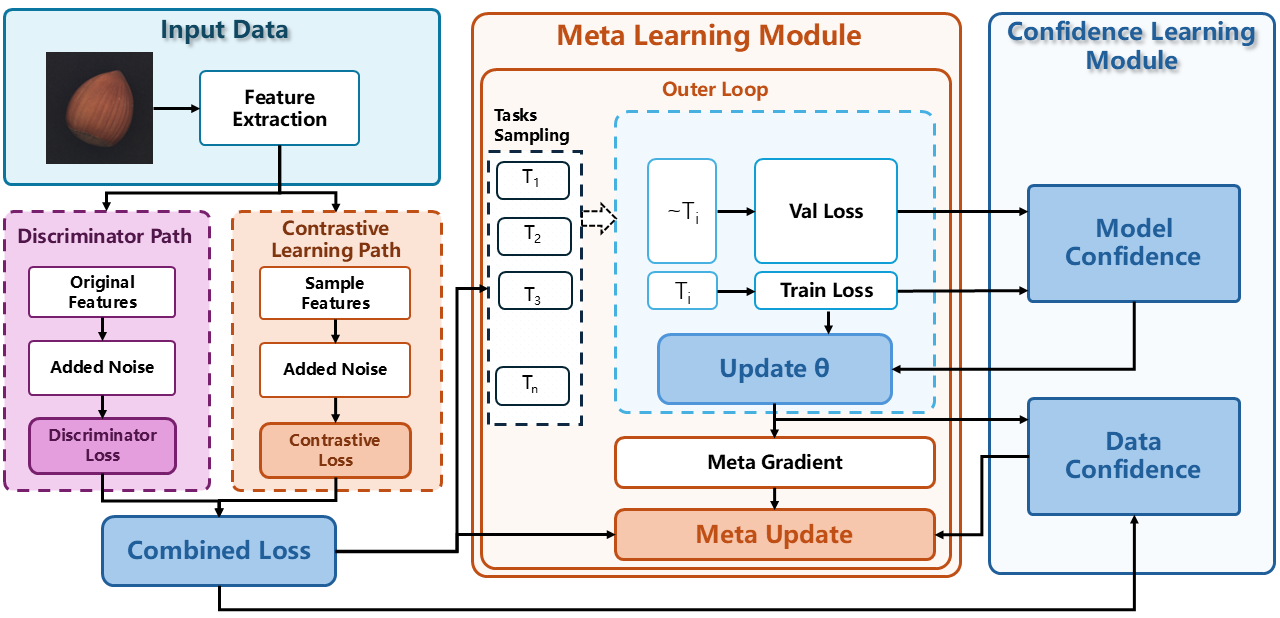}
    \caption{\methname pipeline. A pretrained feature extractor processes input images, feeding into two parallel paths: Discriminator Path (with synthetic anomalies via added noise) and Contrastive Learning Path (with augmented features). Features are divided into tasks for meta-learning, where inner loops optimize individual tasks while outer loops perform meta-updates. The Confident Learning module computes model confidence from training/validation loss covariance and data confidence from anomaly score thresholding, using these to guide parameter updates through adaptive regularization and sample weighting.}
    \label{fig:pipeline}
\end{figure*}

\subsection{Contrastive Learning}

We incorporate patch-based contrastive learning to create a more discriminative feature space for anomaly detection. This approach enhances the model's ability to distinguish normal patterns from anomalies by learning representations where similar normal features cluster together while dissimilar features are pushed apart.

Given feature embeddings $\{x_i\}_{i=1}^{N}$ from normal images, we define the contrastive loss:
\begin{equation}
\resizebox{0.9\columnwidth}{!}{$
\mathcal{L}_{\text{cont}}(x_i) = -\log \frac{\exp(\text{sim}(x_i, \tilde{x}_i) / \tau)}{\exp(\text{sim}(x_i, \tilde{x}_i) / \tau) + \sum_{j \neq i} \exp(\text{sim}(x_i, x_j) / \tau)}
$}
\end{equation}


where $\tilde{x}_i = x_i + \delta$ with $\delta \sim \mathcal{N}(0, \sigma^2 I)$ is an augmented version of $x_i$, $\text{sim}(u, v) = u^T v / \|u\| \|v\|$ is the cosine similarity, and $\tau$ is a temperature parameter.

To enhance learning efficiency, we implement two key innovations:

1. Nearest-neighbor positive mining: For each embedding $x_i$, we identify its $k$ nearest neighbors as additional positive examples. This approach captures the underlying manifold structure of normal data:


\begin{equation}
\resizebox{0.9\columnwidth}{!}{%
$\begin{aligned}
\mathcal{L}_{\text{nn}}(x_i) = -\log \Bigg[ &\frac{\exp(\text{sim}(x_i, \tilde{x}_i) / \tau) + \sum_{m=1}^{k} \exp(\text{sim}(x_i, x_{i,m}) / \tau)}{\exp(\text{sim}(x_i, \tilde{x}_i) / \tau) + \sum_{m=1}^{k} \exp(\text{sim}(x_i, x_{i,m}) / \tau)} \\
&\quad + \sum_{j} \exp(\text{sim}(x_i, x_j) / \tau) \Bigg]
\end{aligned}$%
}
\end{equation}

2. Memory-efficient implementation: We process samples in small chunks to maintain constant memory requirements regardless of dataset size:
\begin{equation}
\mathcal{L}_{\text{batch-cont}} = \frac{1}{B} \sum_{i=1}^{B} \mathcal{L}_{\text{cont}}(x_i)
\end{equation}

The contrastive objective integrates with our confident learning framework:

\begin{equation}
\mathcal{L}_{\text{total}}(\theta) = \mathcal{L}_{\text{SCL}}(\theta) + \lambda_{\text{cont}} \cdot \mathcal{L}_{\text{batch-cont}}
\label{eq:totalloss}
\end{equation}

where $\lambda_{\text{cont}}$ is a weighting coefficient and $\mathcal{L}_{\text{SCL}}(\theta)$ is defined in Equation~\eqref{eq:losscl}. This combined approach creates a feature space where normal patterns form compact clusters, making anomalies more easily identifiable as outliers.

\subsection{Soft Confident Learning}
Our soft confident learning approach aims to differentiate the contribution of training samples based on their representativeness of the normal class distribution. Samples exhibiting strong normal characteristics receive higher influence during training, while those displaying anomalous properties or lying near decision boundaries contribute with reduced impact~\cite{aqeel2025CoMet}. 

This strategy contrasts with conventional confident learning methods that completely discard uncertain samples, as we preserve all data points through a continuous weighting scheme. Operating without access to ground truth labels, our framework simultaneously estimates two distinct uncertainty measures within the unsupervised setting. Sample-level confidence is derived through a combination of model-generated anomaly scores and robust statistical thresholding utilizing interquartile range analysis, providing a probabilistic assessment of each sample's membership in the normal class. The resulting confidence values are incorporated as sample-specific weights within the training objective, creating a hierarchy where prototypical nominal samples strongly influence parameter updates while potentially anomalous samples maintain minimal but non-zero contribution. This approach prevents information loss inherent in hard filtering strategies.

\subsubsection{Data Uncertainty Quantification}
We extend the confident joint principle to unsupervised anomaly detection by analyzing the relationship between training samples and their model-derived confidence assessments.

The model with parameters $\theta$ generates an anomaly score $a_{\theta}(x_i)$ for each sample $x_i$, which inversely correlates with the model's confidence in that sample. Without access to true labels, these scores serve as our primary mechanism for evaluating sample-specific uncertainty. A bounded inverse transformation converts these raw scores into normalized confidence weights:

\begin{equation}
w_i = \min(1, \tau/a_{\theta}(x_i))
\end{equation}

The threshold parameter $\tau$ is computed through robust outlier detection using the Interquartile Range (IQR) methodology~\cite{frery2023interquartile}. Denoting $Q_1$ and $Q_3$ as the first and third quartiles of the anomaly score distribution $a_{\theta}(x)$, we determine:

\begin{equation}
\tau = Q_3 + \kappa \cdot (Q_3 - Q_1)
\label{eq:thiqr}
\end{equation}

where $\kappa$ represents a tunable sensitivity factor adapted to dataset properties.

This weighting mechanism produces our data-uncertainty-aware objective:

\begin{equation}
\mathcal{L}_{data}(\theta) = \sum{i=1}^{N} w_i \cdot \mathcal{L}_{AD}(x_i;\theta)
\end{equation}

with $\mathcal{L}_{AD}(x_i;\theta)$ representing the underlying loss function of the selected anomaly detection architecture.

\subsubsection{Model Uncertainty Quantification}
We capture global model uncertainty by analyzing the geometric properties of the loss landscape through covariance analysis between training and validation losses. This approach leverages statistical theory where the covariance matrix determinant measures multivariate dispersion, with larger values signaling higher uncertainty~\cite{kendall2017uncertainties}. From collected loss sequences $\mathcal{L}{\text{train}}$ and $\mathcal{L}{\text{val}}$ during training iterations, we form the covariance matrix:

\begin{equation}
\Sigma = \begin{bmatrix}
\text{Cov}(\mathcal{L}_{\text{train}}, \mathcal{L}_{\text{train}}) & \text{Cov}(\mathcal{L}_{\text{train}}, \mathcal{L}_{\text{val}}) \\
\text{Cov}(\mathcal{L}_{\text{val}}, \mathcal{L}_{\text{train}}) & \text{Cov}(\mathcal{L}_{\text{val}}, \mathcal{L}_{\text{val}})
\end{bmatrix}
\end{equation}

The determinant $\det(\Sigma)$ yields a scalar uncertainty metric: large values indicate unstable learning with divergent training/validation performance, while small values suggest consistent, confident optimization dynamics.

We integrate this uncertainty measure via an adaptive regularization coefficient that scales dynamically:

\begin{equation}
\lambda( \Sigma ) = \lambda_0 \cdot ( 1 + \gamma \cdot \det(\Sigma) )
\label{eq:model}
\end{equation}

where $\lambda_0$ defines the baseline regularization strength and $\gamma$ modulates sensitivity to uncertainty variations. This mechanism intensifies regularization during high-uncertainty phases to improve generalization while reducing constraints when the model demonstrates stable learning patterns.

Merging sample and model uncertainty components produces our unified soft confident learning objective:

\begin{equation}
\mathcal{L}_{SCL}(\theta) = \sum{i=1}^{N} w_i \cdot \mathcal{L}_{AD}(x_i;\theta) + \lambda(\Sigma) \cdot \Vert \theta \Vert_2^2
\label{eq:losscl}
\end{equation}

\subsection{Meta Learning}
Model-Agnostic Meta-Learning (MAML)~\cite{finn2017model} is incorporated to maintain generalization capabilities when training with heterogeneous sample confidence distributions. Traditional regularization approaches apply uniform parameter constraints, whereas our meta-learning framework provides adaptive, context-sensitive updates essential for varying confidence scenarios. The bilevel optimization structure in MAML, featuring inner task-specific adaptations and outer meta-level updates, delivers three key capabilities.

The framework establishes structured train-validation partitions necessary for our covariance-based uncertainty estimation (Eq.~\ref{eq:totalloss}), creating the mathematical foundation for the adaptive regularization coefficient $\lambda(\Sigma)$. Furthermore, the bilevel architecture automatically discovers parameter updates with strong cross-task generalization properties, implicitly learning effective sample-weighting strategies within the feature representation. Additionally, it maintains balance between confidence-weighted optimization and generalization requirements.

Meta-learning integration facilitates rapid adaptation to distributional shifts while preserving robustness against anomalous patterns. This design achieves equilibrium between task-specific optimization and broad generalization, enabling pattern discovery across varied data distributions while retaining flexibility for domain-specific properties. The meta-learning component ensures that uncertainty-based sample weighting enhances rather than compromises generalization performance.

Task-specific adaptation occurs through inner loop gradient updates, where each task represents a subset of the training data:

\begin{equation}
  \theta' = \theta - \alpha \nabla_{\theta} \mathcal{L}_{\text{train}}(\theta),
\end{equation}

where $\alpha$ denotes the adaptation learning rate and $\mathcal{L}_{\text{train}}(\theta)$ represents the training objective for the current task following equation \eqref{eq:losscl}.

Meta-level optimization updates base parameters using cross-task validation signals, aggregating information from multiple adapted models to improve generalization:

\begin{equation}
  \theta \leftarrow \theta - \beta \nabla_{\theta} \mathcal{L}_{\text{meta}}(\theta'),
\end{equation}

with $\beta$ as the meta-learning rate and $\mathcal{L}_{\text{meta}}(\theta')$ denoting the meta-level objective.

The meta-objective integrates our confidence-weighted formulation:

\begin{equation}
\resizebox{0.8\columnwidth}{!}{$
\mathcal{L}_{\text{meta}}(\theta') = \sum_{i=1}^{N} w_i \cdot \mathcal{L}_{AD}(x_i;\theta') + \lambda(\Sigma') \cdot \Vert \theta' \Vert_2^2,
$}
\end{equation}

where $\theta'$ represents task-adapted parameters, $\mathcal{L}_{AD}(x_i;\theta')$ denotes the sample loss under adapted parameters, $w_i$ indicates recomputed confidence weights using $\theta'$, and $\lambda(\Sigma')$ provides the updated adaptive regularization following equation \eqref{eq:model}. This ensures that sample confidence and model uncertainty adapt throughout the meta-learning process.

\section{Experiment}
\label{sec:experiment}

\subsection{Datasets}
Our ZSAD model underwent rigorous testing across an extensive collection of 10 real-world datasets drawn from both industrial and medical fields. The industrial evaluation leveraged seven established benchmarks: the comprehensive MVTec-AD~\cite{bergmann2019mvtec}, texture-rich VisA~\cite{zou2022spot}, thermal-based BTAD~\cite{mishra2021vt}, manufacturing-focused KSDD2~\cite{bovzivc2021mixed}, rail-specific RSDD~\cite{yu2018coarse}, synthetically-generated DAGM~\cite{fur2007weakly}, and texture-derived DTD-Synthetic~\cite{aota2023zero}. To challenge our model's cross-domain capabilities, we incorporated three medical imaging collections—HeadCT~\cite{salehi2021multiresolution}, BrainMRI~\cite{kanade2015brain}, and Br35H~\cite{Ahamada2020}—each presenting distinct diagnostic challenges. Our methodology employed a strategic approach: we selected VisA as our auxiliary training dataset due to its unique category composition, then evaluated generalization by directly applying the trained model to all other datasets without domain-specific adjustments. For comprehensive validation, we assessed VisA performance separately through models fine-tuned on MVTec-AD, thus enabling thorough examination of our approach's versatility across diverse anomaly detection scenarios.

\subsection{Evaluation Metrics}
We evaluate image-level anomaly detection performance using the standard Area Under the Receiver Operating Characteristic Curve (AUROC), denoted as I-AUROC, based on the anomaly detection score $a_{\theta}(x)$ and equation \eqref{eq:scoreSN}. We also report pixel-level AUROC (P-AUROC).

\subsection{Implementation Details}
We implemented our model using the PyTorch framework and trained it on an NVIDIA RTX 4090 GPU for efficient training and inference. Input images were resized to $256 \times 256$ pixels with optional rotation augmentations. 
 Regularization was achieved through weight decay to prevent overfitting. Training was conducted over 40 epochs with a batch size of 16 and a learning rate of $2 \times 10^{-4}$. This setup enables our model to effectively learn and generalize complex data distributions, achieving robust performance in density estimation and generative tasks.

\begin{table*}[t!]
\centering
\caption{Comparison of Image-level AUROC (I-AUROC) values with existing state-of-the-art methods. The best results are marked with \textbf{Bold} text and darker blue background, second-best with medium blue background, and third-best with lighter blue background.}
\label{table:i_auroc_comparison}
\resizebox{\textwidth}{!}{%
\begin{tabular}{ccccccccc}
\hline
\makecell{Domain} & \makecell{Dataset} & \makecell{AnomalyCLIP\\~\cite{zhou2023anomalyclip}} & \makecell{APRIL-GAN\\~\cite{chen2023april}} & \makecell{AdaCLIP\\~\cite{cao2024adaclip}} & \makecell{WinCLIP\\~\cite{jeong2023winclip}} & \makecell{CLIP-AD\\~\cite{chen2024clip}} & \makecell{Bayes-PFL\\~\cite{qu2025bayesian}} & \makecell{\textbf{\emph{\methname}}} \\[8pt]
\hline
\multirow{7}{*}{Industrial} 
& MVTec-AD & 91.5 & 86.1 & 92.0 & 91.8 & 89.8 & \cellcolor{second}92.3 & \cellcolor{first}\textbf{92.9}\\
& VisA & 82.1 & 78.0 & \cellcolor{third}83.0 & 78.1 & 79.8 & \cellcolor{first}\textbf{87.0} & \cellcolor{second}85.3 \\
& BTAD & 89.1 & 74.2 & \cellcolor{third}91.6 & 83.3 & 85.8 & \cellcolor{second}93.2 & \cellcolor{first}\textbf{97.2} \\
& KSDD2 & 92.1 & 90.3 & \cellcolor{third}95.9 & 93.5 & 95.2 & \cellcolor{first}\textbf{97.2} & \cellcolor{second}94.2 \\
& RSDD & 73.5 & 73.1 & \cellcolor{third}89.1 & 85.3 & 88.3 & \cellcolor{second}94.1 & \cellcolor{first}\textbf{94.7}\\
& DAGM & 95.6 & 90.4 & \cellcolor{third}96.5 & 89.6 & 90.8 & \cellcolor{second}97.7 & \cellcolor{first}\textbf{97.9} \\
& DTD-Synthetic & 94.5 & 83.9 & 92.8 & \cellcolor{third}95.0 & 91.5 & \cellcolor{second}95.1 & \cellcolor{first}\textbf{99.2} \\
\hline
\multirow{3}{*}{Medical} 
& HeadCT & \cellcolor{third}95.3 & 89.3 & 93.4 & 83.7 & 93.8 & \cellcolor{second}96.5 & \cellcolor{first}\textbf{96.8}\\
& BrainMRI & \cellcolor{third}96.1 & 89.6 & 94.9 & 92.0 & 92.8 & \cellcolor{second}\textbf{96.2} & \cellcolor{first}\textbf{98.7} \\
& Br35H & \cellcolor{third}97.3 & 93.1 & 95.7 & 80.5 & 96.0 & \cellcolor{second}97.8 & \cellcolor{first}\textbf{99.3}\\
\hline
\end{tabular}%
}
\end{table*}

\begin{table*}[htbp]
\centering
\caption{Comparison of Pixel-level AUROC (P-AUROC) values with existing state-of-the-art methods. The best results are marked with \textbf{Bold} text and darker blue background, second-best with medium blue background, and third-best with lighter blue background.}
\label{table:p_auroc_comparison}
\resizebox{\textwidth}{!}{%
\begin{tabular}{ccccccccc}
\hline
\makecell{Domain} & \makecell{Dataset} & \makecell{AnomalyCLIP\\~\cite{zhou2023anomalyclip}} & \makecell{APRIL-GAN\\~\cite{chen2023april}} & \makecell{AdaCLIP\\~\cite{cao2024adaclip}} & \makecell{WinCLIP\\~\cite{jeong2023winclip}} & \makecell{CLIP-AD\\~\cite{chen2024clip}} & \makecell{Bayes-PFL\\~\cite{qu2025bayesian}} & \makecell{\textbf{\emph{\methname}}} \\[8pt]
\hline
\multirow{7}{*}{Industrial} 
& MVTec-AD & 91.1 & 87.6 & 86.8 & 85.1 & 89.8 & \cellcolor{second}91.8 & \cellcolor{first}\textbf{96.3} \\
& VisA & \cellcolor{third}95.5 & 94.2 & 95.1 & 79.6 & 95.0 & \cellcolor{second}95.6 & \cellcolor{first}\textbf{95.9} \\
& BTAD & \cellcolor{second}93.3 & 91.3 & 87.7 & 71.4 & 93.1 & \cellcolor{first}\textbf{93.9} & \cellcolor{third}93.5 \\
& KSDD2 & \cellcolor{second}99.4 & 97.9 & 96.1 & 97.9 & \cellcolor{third}99.3 & \cellcolor{first}\textbf{99.6} & 97.8 \\
& RSDD & 99.1 & 99.4 & \cellcolor{third}99.5 & 95.1 & 99.2 & \cellcolor{second}99.6 & \cellcolor{first}\textbf{99.7} \\
& DAGM & 99.1 & \cellcolor{third}99.2 & 97.0 & 83.2 & 99.0 & \cellcolor{second}99.3 & \cellcolor{first}\textbf{99.5} \\
& DTD-Synthetic & \cellcolor{third}97.6 & 96.6 & 94.1 & 82.5 & 97.1 & \cellcolor{second}97.8 & \cellcolor{first}\textbf{97.9}\\
\hline
\multirow{3}{*}{Medical} 
& HeadCT & \cellcolor{third}82.6 & 75.8 & 82.2 & 73.3 & 81.4 & \cellcolor{second}87.4 & \cellcolor{first}\textbf{94.3}\\
& BrainMRI & \cellcolor{second}88.4 & \cellcolor{third}85.8 & 85.4 & 83.3 & 81.6 & \cellcolor{first}92.2 & 84.9 \\
& BR35H & 51.9 & 58.4 & 59.3 & \cellcolor{second}64.8 & \cellcolor{third}60.3 & 57.1 & \cellcolor{first}\textbf{79.9}\\
\hline
\end{tabular}%
}
\end{table*}

\subsection{\methname Model Configuration}
\label{sec:zsad}
\emph{\methname} employs a modified SimpleNet~\cite{liu2023simplenet} architecture for zero-shot anomaly detection. The backbone consists of a WideResNet-50 pre-trained on ImageNet to extract rich feature representations from input images $x_i$, producing local features $o_i=F_*(x_i)$, where the $*$ symbol indicates that the feature extractor parameters are frozen. We utilize the 2nd and 3rd intermediate layers of the backbone with a total feature dimension of 1536.

A feature adaptor $G_{\theta_1}$, implemented as a single fully-connected layer without bias, transforms the extracted features to a task-specific representation space:
\begin{equation}
    q_i = G_{\theta_1}(o_i) = G_{\theta_1}(F_*(x_i))
    \label{eq:feature_adaptor}
\end{equation}

During training, synthetic anomalies are generated by adding Gaussian noise to normal features:
\begin{equation}
    q_i^- = q_i + \epsilon, \; \epsilon \sim\mathcal{N}(0,\sigma^2)
    \label{eq:synthetic_anomaly}
\end{equation}
where $\sigma=0.015$ controls the perturbation intensity. This simple yet effective approach allows the model to learn anomaly characteristics without requiring actual anomalous samples.

The discriminator $D_{\theta_2}$ consists of a linear layer, batch normalization, LeakyReLU activation (0.2 slope), and a final linear layer. It acts as a normality scorer that estimates whether features correspond to normal or anomalous regions. The discriminator is trained using a margin-based loss function:
\begin{equation}
    \resizebox{0.8\columnwidth}{!}{$
    \mathcal{L}_{AD}(x|\theta) = \max(0,th_+ - D_{\theta_2}(q_i)) + \max(0, D_{\theta_2}(q_i^-) + th_-)
    $}
    \label{eq:loss_ad}
\end{equation}
where $th_+$ and $th_-$ are saturation thresholds both set to 0.5, and $\theta=\{\theta_1,\theta_2\}$ represents the complete model parameters for both the feature adaptor and discriminator. The model is optimized using Adam with learning rates of 0.0001 for the feature adaptor and 0.0002 for the discriminator, with weight decay set to 0.00001. 

During inference, the anomaly feature generator is discarded, and the discriminator directly computes the anomaly score:
\begin{equation}
    a_{\theta}(x_i) = -D_{\theta_2}(G_{\theta_1}(F_*(x_i)))
    \label{eq:scoreSN}
\end{equation}

This approach enables effective zero-shot anomaly detection by learning to distinguish normal patterns without requiring any anomalous samples during training.

\subsection{Results}

\emph{\methname} demonstrates exceptional performance on industrial datasets, achieving state-of-the-art results on 6 out of 7 benchmarks (Table~\ref{table:i_auroc_comparison}). Our method excels particularly in texture anomaly detection, with substantial improvements of 4.1\% on DTD-Synthetic (99.2\% I-AUROC) and 4.0\% on BTAD (97.2\% I-AUROC) compared to previous best methods. On MVTec-AD, \emph{\methname} achieves 92.9\% I-AUROC, representing a 0.6\% improvement over Bayes-PFL that is significant given existing methods approach optimal performance on this benchmark. Our method also sets new benchmarks on RSDD (94.7\%) and DAGM (97.9\%), confirming robust performance across diverse industrial scenarios.

For pixel-level detection (Table~\ref{table:p_auroc_comparison}), \emph{\methname} establishes new state-of-the-art performance on most datasets, with particularly notable localization improvements. On MVTec-AD, we achieve 96.3\% P-AUROC, outperforming the previous best by 4.5\%. This substantial enhancement in boundary delineation highlights our method's precision in identifying exact anomaly regions. Our approach maintains strong localization performance, achieving the highest P-AUROC on VisA (95.9\%), RSDD (99.7\%), DAGM (99.5\%), and DTD-Synthetic (97.9\%). However, on KSDD2, our method achieves 97.8\% P-AUROC, ranking fourth behind methods exceeding 99\%, suggesting opportunities for boundary refinement in specific defect types.

\emph{\methname} demonstrates strong transferability to medical imaging domains without domain-specific retraining. On HeadCT, our method achieves the highest performance with 96.8\% I-AUROC and 94.3\% P-AUROC, with localization improvements of 6.9\% over the previous best method. For BrainMRI and Br35H, \emph{\methname} achieves 98.7\% and 99.3\% I-AUROC respectively, outperforming all competing methods and confirming our framework's ability to generalize across different medical imaging modalities.

Our comprehensive evaluation reveals that \emph{\methname} consistently outperforms existing approaches in most scenarios, with particular strengths in texture-based defect detection and precise anomaly localization. The framework's ability to maintain high performance across diverse domains validates our soft confident learning approach that preserves boundary information while focusing on prototypical patterns. We attribute this strong performance to the effective integration of confident learning, meta-learning, and contrastive representation, enabling robust anomaly detection while maintaining adaptability across varying conditions and defect types.

\subsection{Ablation Study}

We conduct comprehensive ablation experiments to validate the effectiveness of each component in our \emph{\methname} framework. All experiments are performed on MVTec-AD and VisA datasets with consistent hyperparameters unless stated otherwise.

\subsubsection{Component Analysis}

Table~\ref{table:component_ablation} demonstrates the contribution of each major component. The baseline SimpleNet achieves $89.7\%$ I-AUROC on MVTec-AD, while progressively adding components shows cumulative improvements. Confident Learning provides the most significant boost ($+2.1\%$), followed by Meta-Learning ($+0.8\%$) and Contrastive Learning ($+0.3\%$). The complete \emph{\methname} framework achieves $92.9\%$ I-AUROC, representing a $3.2\%$ improvement over the baseline.

On VisA, the pattern differs with Meta-Learning showing the largest contribution ($+3.1\%$), highlighting its importance for cross-domain generalization. Confident Learning contributes $+1.8\%$, while Contrastive Learning adds $+0.4\%$. This variation across datasets demonstrates that different components address different challenges in anomaly detection.

\begin{table}[h]
\centering
\caption{Component-wise ablation study showing I-AUROC (\%).}
\label{table:component_ablation}
\begin{tabular}{lcc}
\hline
Method & MVTec-AD & VisA \\
\hline
Baseline (SimpleNet) & 89.7 & 80.0 \\
+ Confident Learning & 91.8 & 81.8 \\
+ Meta-Learning & 92.6 & 84.9 \\
+ Contrastive Learning (\textbf{\emph{\methname}}) & \textbf{92.9} & \textbf{85.3} \\
\hline
\end{tabular}
\end{table}

\subsubsection{Confident Learning Module Analysis}

We analyze the individual contributions of data uncertainty weighting and model uncertainty regularization within our Confident Learning module (Table~\ref{table:cl_ablation}). Removing data uncertainty weighting (w/o Data Uncertainty) shows larger performance drops than removing model uncertainty regularization (w/o Model Uncertainty), indicating that sample-level confidence scoring is more critical than adaptive regularization.

Comparing hard filtering versus our soft weighting approach reveals that completely removing uncertain samples (Hard CL) performs worse than proportionally down-weighting them. This validates our design choice of preserving all training data while modulating their influence based on confidence.

\begin{table}[h]
\centering
\caption{Confident Learning module ablation showing I-AUROC (\%).}
\label{table:cl_ablation}
\begin{tabular}{lcc}
\hline
Method & MVTec-AD & VisA \\
\hline
\textbf{Full \emph{\methname}} & \textbf{92.9} & \textbf{85.3} \\
w/o Data Uncertainty & 91.4 & 83.7 \\
w/o Model Uncertainty & 92.1 & 84.5 \\
w/o Confident Learning & 90.8 & 81.2 \\
Hard CL (remove samples) & 91.6 & 83.1 \\
\hline
\end{tabular}
\end{table}

\subsubsection{Parameter Sensitivity Analysis}

Table~\ref{table:parameter_sensitivity} examines the sensitivity of key hyperparameters. The IQR threshold parameter $\kappa$ shows optimal performance at $\kappa=1.5$, with degradation at both lower and higher values. Values below $1.0$ include too many uncertain samples, while values above $2.0$ remove potentially informative data.

For nearest-neighbor mining in contrastive learning, $K_{nn}=5$ provides the best balance between capturing local manifold structure and computational efficiency. Lower $K_{nn}$ values fail to capture sufficient positive relationships, while higher values introduce noise from distant neighbors.

The contrastive loss weight $\lambda_{\text{cont}} = 1.0$ achieves optimal integration with the confident learning objective. Lower weights underutilize contrastive learning benefits, while higher weights overwhelm the primary anomaly detection loss.

\begin{table}[h]
\centering
\caption{Parameter sensitivity analysis showing I-AUROC (\%) on MVTec-AD.}
\label{table:parameter_sensitivity}
\begin{tabular}{lcc}
\hline
Parameter & Value & I-AUROC \\
\hline
\multirow{4}{*}{$\kappa$ (IQR)} 
 & 0.5 & 91.8 \\
 & 1.0 & 92.3 \\
 & 1.5 & \textbf{92.9} \\
 & 2.0 & 92.1 \\
\hline
\multirow{4}{*}{$k_{nn}$}
 & 1 & 92.1 \\
 & 3 & 92.4 \\
 & 5 & \textbf{92.9} \\
 & 10 & 92.3 \\
\hline
\multirow{4}{*}{$\lambda_{\text{cont}}$}
 & 0.1 & 92.2 \\
 & 0.5 & 92.6 \\
 & 1.0 & \textbf{92.9} \\
 & 2.0 & 92.4 \\
\hline
\end{tabular}
\end{table}

\subsubsection{Feature Layer Analysis}

We investigate the impact of using different combinations of intermediate layers from the WideResNet-50 backbone (Table~\ref{table:layer_ablation}). The combination of 2nd and 3rd layers (our default choice) provides the best trade-off between semantic richness and spatial resolution. Earlier layers (1st+2nd) lack sufficient semantic information, while later layers (3rd+4th) lose fine-grained spatial details crucial for precise anomaly localization.

\begin{table}[h]
\centering
\caption{Feature layer combination analysis showing I-AUROC/P-AUROC (\%).}
\label{table:layer_ablation}
\begin{tabular}{lcc}
\hline
Layer Combination & MVTec-AD & VisA \\
\hline
1st + 2nd layers & 91.2/94.8 & 83.1/94.2 \\
2nd + 3rd layers & \textbf{92.9/96.3} & \textbf{85.3/95.9} \\
3rd + 4th layers & 92.1/95.1 & 84.7/95.3 \\
All layers (1st-4th) & 92.3/95.7 & 84.9/95.6 \\
\hline
\end{tabular}
\end{table}

\subsubsection{Component Interaction Analysis}

To understand synergistic effects, we evaluate pairwise component combinations (Table~\ref{table:interaction_ablation}). The combination of Soft Confident Learning with Meta-Learning (SCL+Meta) achieves substantial improvements over individual components, particularly on VisA ($+4.9\%$ vs baseline). Adding Contrastive Learning provides smaller but consistent gains, suggesting diminishing returns when all uncertainty and adaptation mechanisms are already in place.

\begin{table}[h]
\centering
\caption{Component interaction analysis showing I-AUROC (\%).}
\label{table:interaction_ablation}
\begin{tabular}{lcc}
\hline
Method & MVTec-AD & VisA \\
\hline
Baseline & 89.7 & 80.0 \\
SCL only & 91.8 & 81.8 \\
Meta only & 90.5 & 83.1 \\
Contrastive only & 90.1 & 80.4 \\
SCL + Meta & 92.6 & 84.9 \\
SCL + Contrastive & 92.0 & 82.1 \\
Meta + Contrastive & 90.8 & 83.4 \\
\textbf{Full \emph{\methname}} & \textbf{92.9} & \textbf{85.3} \\
\hline
\end{tabular}
\end{table}

The ablation studies confirm that each component contributes meaningfully to overall performance, with Confident Learning providing the most consistent improvements and Meta-Learning being crucial for cross-domain generalization. The framework demonstrates robustness to hyperparameter choices while achieving optimal performance with our selected configuration.

\section{Conclusion}
\label{sec:conclusion}

We presented \emph{\methname}, a novel zero-shot anomaly detection framework that integrates soft confident learning, meta-learning, and contrastive feature representation to address critical limitations in industrial and medical anomaly detection. Unlike traditional confident learning approaches that discard uncertain samples, our soft weighting mechanism preserves valuable boundary information while emphasizing prototypical normal patterns, enabling more robust detection without requiring anomalous training data. The framework quantifies both data and model uncertainty through statistical thresholding and covariance-based metrics, creating a principled approach to sample weighting that enhances learning from nominal data alone.
Our comprehensive evaluation across 10 diverse datasets demonstrates that \emph{\methname} achieves state-of-the-art performance across industrial and medical domains, with particularly strong results on texture-rich datasets and superior pixel-level localization capabilities. The ablation studies validate each component's contribution, confirming that confident learning provides consistent improvements while meta-learning enables effective cross-domain generalization. The framework's robustness to hyperparameter choices and ability to transfer knowledge across different anomaly detection scenarios establishes its practical viability for real-world deployment where labeled anomalous data is scarce or unavailable.

\section{Future Work and Limitations}
\label{sec:futurework}

While \emph{\methname} demonstrates strong performance across diverse domains, several limitations present opportunities for future research. The framework's reliance on IQR-based thresholding may not optimally adapt to all data distributions, suggesting potential benefits from learnable threshold mechanisms. Computational efficiency for real-time industrial deployment remains an area for optimization, particularly for high-resolution imagery. Future work should explore integration with explainable AI techniques to provide interpretable anomaly explanations crucial for medical applications, investigation of few-shot adaptation mechanisms for rapid domain customization, and extension to additional modalities including video sequences and 3D data. Additionally, federated learning approaches could enable collaborative model improvement across institutions while preserving data privacy, particularly valuable in healthcare settings.

\section*{Acknowledgements}

This study was carried out within the PNRR research activities of the
consortium iNEST (Interconnected North-Est Innovation Ecosystem) funded by the European Union Next-GenerationEU (Piano Nazionale di Ripresa e Resilienza (PNRR) – Missione 4 Componente 2, Investimento 1.5 – D.D. 1058  23/06/2022, ECS\_00000043).

{
    \small
    \bibliographystyle{ieeenat_fullname}
    \bibliography{main}

\begin{thebibliography}{40}
\providecommand{\natexlab}[1]{#1}
\providecommand{\url}[1]{\texttt{#1}}
\expandafter\ifx\csname urlstyle\endcsname\relax
  \providecommand{\doi}[1]{doi: #1}\else
  \providecommand{\doi}{doi: \begingroup \urlstyle{rm}\Url}\fi

\bibitem[Aota et~al.(2023)Aota, Tong, and Okatani]{aota2023zero}
Toshimichi Aota, Lloyd Teh~Tzer Tong, and Takayuki Okatani.
\newblock Zero-shot versus many-shot: Unsupervised texture anomaly detection.
\newblock In \emph{Proceedings of the IEEE/CVF Winter Conference on Applications of Computer Vision}, pages 5564--5572, 2023.

\bibitem[Aqeel et~al.(2025{\natexlab{a}})Aqeel, Sharifi, Cristani, and Setti]{aqeel2025CoMet}
Muhammad Aqeel, Shakiba Sharifi, Marco Cristani, and Francesco Setti.
\newblock Towards real unsupervised anomaly detection via confident meta-learning.
\newblock In \emph{Proceedings of the ieee/cvf international conference on computer vision}, 2025{\natexlab{a}}.

\bibitem[Aqeel et~al.(2025{\natexlab{b}})Aqeel, Sharifi, Cristani, and Setti]{aqeel2025meta}
Muhammad Aqeel, Shakiba Sharifi, Marco Cristani, and Francesco Setti.
\newblock Meta learning-driven iterative refinement for robust anomaly detection in industrial inspection.
\newblock In \emph{European Conference on Computer Vision}, pages 445--460. Springer, 2025{\natexlab{b}}.

\bibitem[Bao et~al.(2024)Bao, Sun, Deng, He, Zhang, and Li]{bao2024bmad}
Jinan Bao, Hanshi Sun, Hanqiu Deng, Yinsheng He, Zhaoxiang Zhang, and Xingyu Li.
\newblock Bmad: Benchmarks for medical anomaly detection.
\newblock In \emph{Proceedings of the IEEE/CVF Conference on Computer Vision and Pattern Recognition}, pages 4042--4053, 2024.

\bibitem[Baugh et~al.(2023)Baugh, Batten, M{\"u}ller, and Kainz]{baugh2023zero}
Matthew Baugh, James Batten, Johanna~P M{\"u}ller, and Bernhard Kainz.
\newblock Zero-shot anomaly detection with pre-trained segmentation models.
\newblock \emph{arXiv preprint arXiv:2306.09269}, 2023.

\bibitem[Bergmann et~al.(2019)Bergmann, Fauser, Sattlegger, and Steger]{bergmann2019mvtec}
Paul Bergmann, Michael Fauser, David Sattlegger, and Carsten Steger.
\newblock Mvtec ad--a comprehensive real-world dataset for unsupervised anomaly detection.
\newblock In \emph{Proceedings of the IEEE/CVF conference on computer vision and pattern recognition}, pages 9592--9600, 2019.

\bibitem[Bergmann et~al.(2020)Bergmann, Fauser, Sattlegger, and Steger]{bergmann2020uninformed}
Paul Bergmann, Michael Fauser, David Sattlegger, and Carsten Steger.
\newblock Uninformed students: Student-teacher anomaly detection with discriminative latent embeddings.
\newblock In \emph{Proceedings of the IEEE/CVF conference on computer vision and pattern recognition}, pages 4183--4192, 2020.

\bibitem[Bo{\v{z}}i{\v{c}} et~al.(2021)Bo{\v{z}}i{\v{c}}, Tabernik, and Sko{\v{c}}aj]{bovzivc2021mixed}
Jakob Bo{\v{z}}i{\v{c}}, Domen Tabernik, and Danijel Sko{\v{c}}aj.
\newblock Mixed supervision for surface-defect detection: From weakly to fully supervised learning.
\newblock \emph{Computers in Industry}, 129:\penalty0 103459, 2021.

\bibitem[Cao et~al.(2023)Cao, Xu, Sun, Cheng, Du, Gao, and Shen]{cao2023segment}
Yunkang Cao, Xiaohao Xu, Chen Sun, Yuqi Cheng, Zongwei Du, Liang Gao, and Weiming Shen.
\newblock Segment any anomaly without training via hybrid prompt regularization.
\newblock \emph{arXiv preprint arXiv:2305.10724}, 2023.

\bibitem[Cao et~al.(2024)Cao, Zhang, Frittoli, Cheng, Shen, and Boracchi]{cao2024adaclip}
Yunkang Cao, Jiangning Zhang, Luca Frittoli, Yuqi Cheng, Weiming Shen, and Giacomo Boracchi.
\newblock Adaclip: Adapting clip with hybrid learnable prompts for zero-shot anomaly detection.
\newblock In \emph{European Conference on Computer Vision}, pages 55--72. Springer, 2024.

\bibitem[Chen et~al.(2024{\natexlab{a}})Chen, Luo, Lv, and Zhang]{chen2024unified}
Qiyu Chen, Huiyuan Luo, Chengkan Lv, and Zhengtao Zhang.
\newblock A unified anomaly synthesis strategy with gradient ascent for industrial anomaly detection and localization.
\newblock In \emph{European Conference on Computer Vision}, pages 37--54. Springer, 2024{\natexlab{a}}.

\bibitem[Chen et~al.(2023)Chen, Han, and Zhang]{chen2023april}
Xuhai Chen, Yue Han, and Jiangning Zhang.
\newblock April-gan: A zero-/few-shot anomaly classification and segmentation method for cvpr 2023 vand workshop challenge tracks 1\&2: 1st place on zero-shot ad and 4th place on few-shot ad.
\newblock \emph{arXiv preprint arXiv:2305.17382}, 2023.

\bibitem[Chen et~al.(2024{\natexlab{b}})Chen, Zhang, Tian, He, Zhang, Wang, Wang, and Liu]{chen2024clip}
Xuhai Chen, Jiangning Zhang, Guanzhong Tian, Haoyang He, Wuhao Zhang, Yabiao Wang, Chengjie Wang, and Yong Liu.
\newblock Clip-ad: A language-guided staged dual-path model for zero-shot anomaly detection.
\newblock In \emph{International Joint Conference on Artificial Intelligence}, pages 17--33. Springer, 2024{\natexlab{b}}.

\bibitem[Finn et~al.(2017)Finn, Abbeel, and Levine]{finn2017model}
Chelsea Finn, Pieter Abbeel, and Sergey Levine.
\newblock Model-agnostic meta-learning for fast adaptation of deep networks.
\newblock In \emph{International Conference on Machine Learning (ICML)}, pages 1126--1135, 2017.

\bibitem[Frery(2023)]{frery2023interquartile}
Alejandro~C Frery.
\newblock Interquartile range.
\newblock In \emph{Encyclopedia of Mathematical Geosciences}, pages 664--666. Springer, 2023.

\bibitem[f{\"u}r Mustererkennung(2007)]{fur2007weakly}
Deutsche~Arbeitsgemeinschaft f{\"u}r Mustererkennung.
\newblock Weakly supervised learning for industrial optical inspection, 2007.

\bibitem[Gu et~al.(2024)Gu, Zhu, Zhu, Chen, Li, Tang, and Wang]{gu2024filo}
Zhaopeng Gu, Bingke Zhu, Guibo Zhu, Yingying Chen, Hao Li, Ming Tang, and Jinqiao Wang.
\newblock Filo: Zero-shot anomaly detection by fine-grained description and high-quality localization.
\newblock In \emph{Proceedings of the 32nd ACM International Conference on Multimedia}, pages 2041--2049, 2024.

\bibitem[Hamada(2020)]{Ahamada2020}
Ahmed Hamada.
\newblock Br35h: Brain tumor detection 2020.
\newblock 2020.

\bibitem[He et~al.(2024)He, Jiang, Peng, Zhu, Liu, Du, Hu, Chi, Wang, and Wang]{he2024learning}
Liren He, Zhengkai Jiang, Jinlong Peng, Wenbing Zhu, Liang Liu, Qiangang Du, Xiaobin Hu, Mingmin Chi, Yabiao Wang, and Chengjie Wang.
\newblock Learning unified reference representation for unsupervised multi-class anomaly detection.
\newblock In \emph{European Conference on Computer Vision}, pages 216--232. Springer, 2024.

\bibitem[Huang et~al.(2022)Huang, Guan, Jiang, Zhang, Spratling, and Wang]{huang2022registration}
Chaoqin Huang, Haoyan Guan, Aofan Jiang, Ya Zhang, Michael Spratling, and Yan-Feng Wang.
\newblock Registration based few-shot anomaly detection.
\newblock In \emph{European conference on computer vision}, pages 303--319. Springer, 2022.

\bibitem[Huang et~al.(2024)Huang, Jiang, Feng, Zhang, Wang, and Wang]{huang2024adapting}
Chaoqin Huang, Aofan Jiang, Jinghao Feng, Ya Zhang, Xinchao Wang, and Yanfeng Wang.
\newblock Adapting visual-language models for generalizable anomaly detection in medical images.
\newblock In \emph{Proceedings of the IEEE/CVF Conference on Computer Vision and Pattern Recognition}, pages 11375--11385, 2024.

\bibitem[Jeong et~al.(2023)Jeong, Zou, Kim, Zhang, Ravichandran, and Dabeer]{jeong2023winclip}
Jongheon Jeong, Yang Zou, Taewan Kim, Dongqing Zhang, Avinash Ravichandran, and Onkar Dabeer.
\newblock Winclip: Zero-/few-shot anomaly classification and segmentation.
\newblock In \emph{Proceedings of the IEEE/CVF Conference on Computer Vision and Pattern Recognition}, pages 19606--19616, 2023.

\bibitem[Kanade and Gumaste(2015)]{kanade2015brain}
Pranita~Balaji Kanade and PP Gumaste.
\newblock Brain tumor detection using mri images.
\newblock \emph{Brain}, 3\penalty0 (2):\penalty0 146--150, 2015.

\bibitem[Kendall and Gal(2017)]{kendall2017uncertainties}
Alex Kendall and Yarin Gal.
\newblock What uncertainties do we need in bayesian deep learning for computer vision?
\newblock \emph{Advances in neural information processing systems}, 30, 2017.

\bibitem[Kirillov et~al.(2023)Kirillov, Mintun, Ravi, Mao, Rolland, Gustafson, Xiao, Whitehead, Berg, Lo, et~al.]{kirillov2023segment}
Alexander Kirillov, Eric Mintun, Nikhila Ravi, Hanzi Mao, Chloe Rolland, Laura Gustafson, Tete Xiao, Spencer Whitehead, Alexander~C Berg, Wan-Yen Lo, et~al.
\newblock Segment anything.
\newblock In \emph{Proceedings of the IEEE/CVF international conference on computer vision}, pages 4015--4026, 2023.

\bibitem[Li et~al.(2025)Li, Cao, Ye, Ding, Tu, and Chen]{li2025clipsam}
Shengze Li, Jianjian Cao, Peng Ye, Yuhan Ding, Chongjun Tu, and Tao Chen.
\newblock Clipsam: Clip and sam collaboration for zero-shot anomaly segmentation.
\newblock \emph{Neurocomputing}, 618:\penalty0 129122, 2025.

\bibitem[Liu et~al.(2024)Liu, Zeng, Ren, Li, Zhang, Yang, Jiang, Li, Yang, Su, et~al.]{liu2024grounding}
Shilong Liu, Zhaoyang Zeng, Tianhe Ren, Feng Li, Hao Zhang, Jie Yang, Qing Jiang, Chunyuan Li, Jianwei Yang, Hang Su, et~al.
\newblock Grounding dino: Marrying dino with grounded pre-training for open-set object detection.
\newblock In \emph{European Conference on Computer Vision}, pages 38--55. Springer, 2024.

\bibitem[Liu et~al.(2023)Liu, Zhou, Xu, and Wang]{liu2023simplenet}
Zhikang Liu, Yiming Zhou, Yuansheng Xu, and Zilei Wang.
\newblock Simplenet: A simple network for image anomaly detection and localization.
\newblock In \emph{IEEE/CVF Conference on Computer Vision and Pattern Recognition (CVPR)}, pages 20402--20411, 2023.

\bibitem[Mishra et~al.(2021)Mishra, Verk, Fornasier, Piciarelli, and Foresti]{mishra2021vt}
Pankaj Mishra, Riccardo Verk, Daniele Fornasier, Claudio Piciarelli, and Gian~Luca Foresti.
\newblock Vt-adl: A vision transformer network for image anomaly detection and localization.
\newblock In \emph{2021 IEEE 30th International Symposium on Industrial Electronics (ISIE)}, pages 01--06. IEEE, 2021.

\bibitem[Qu et~al.(2024)Qu, Tao, Prasad, Shen, Zhang, Gong, and Ding]{qu2024vcp}
Zhen Qu, Xian Tao, Mukesh Prasad, Fei Shen, Zhengtao Zhang, Xinyi Gong, and Guiguang Ding.
\newblock Vcp-clip: A visual context prompting model for zero-shot anomaly segmentation.
\newblock In \emph{European Conference on Computer Vision}, pages 301--317. Springer, 2024.

\bibitem[Qu et~al.(2025)Qu, Tao, Gong, Qu, Chen, Zhang, Wang, and Ding]{qu2025bayesian}
Zhen Qu, Xian Tao, Xinyi Gong, Shichen Qu, Qiyu Chen, Zhengtao Zhang, Xingang Wang, and Guiguang Ding.
\newblock Bayesian prompt flow learning for zero-shot anomaly detection.
\newblock \emph{arXiv preprint arXiv:2503.10080}, 2025.

\bibitem[Salehi et~al.(2021)Salehi, Sadjadi, Baselizadeh, Rohban, and Rabiee]{salehi2021multiresolution}
Mohammadreza Salehi, Niousha Sadjadi, Soroosh Baselizadeh, Mohammad~H Rohban, and Hamid~R Rabiee.
\newblock Multiresolution knowledge distillation for anomaly detection.
\newblock In \emph{Proceedings of the IEEE/CVF conference on computer vision and pattern recognition}, pages 14902--14912, 2021.

\bibitem[Xu et~al.(2025)Xu, Lo, Safaei, Patel, and Dwivedi]{xu2025towards}
Jiacong Xu, Shao-Yuan Lo, Bardia Safaei, Vishal~M Patel, and Isht Dwivedi.
\newblock Towards zero-shot anomaly detection and reasoning with multimodal large language models.
\newblock \emph{arXiv preprint arXiv:2502.07601}, 2025.

\bibitem[Yao et~al.(2024)Yao, Liu, Yin, Yan, Hong, and Zuo]{yao2024glad}
Hang Yao, Ming Liu, Zhicun Yin, Zifei Yan, Xiaopeng Hong, and Wangmeng Zuo.
\newblock Glad: towards better reconstruction with global and local adaptive diffusion models for unsupervised anomaly detection.
\newblock In \emph{European Conference on Computer Vision}, pages 1--17. Springer, 2024.

\bibitem[You et~al.(2022)You, Cui, Shen, Yang, Lu, Zheng, and Le]{you2022unified}
Zhiyuan You, Lei Cui, Yujun Shen, Kai Yang, Xin Lu, Yu Zheng, and Xinyi Le.
\newblock A unified model for multi-class anomaly detection.
\newblock \emph{Advances in Neural Information Processing Systems}, 35:\penalty0 4571--4584, 2022.

\bibitem[Yu et~al.(2018)Yu, Li, Tan, Gan, Wang, Geng, and Jia]{yu2018coarse}
Haomin Yu, Qingyong Li, Yunqiang Tan, Jinrui Gan, Jianzhu Wang, Yangli-ao Geng, and Lei Jia.
\newblock A coarse-to-fine model for rail surface defect detection.
\newblock \emph{IEEE Transactions on Instrumentation and Measurement}, 68\penalty0 (3):\penalty0 656--666, 2018.

\bibitem[Zhao et~al.(2021)Zhao, Li, He, Ma, Fang, Li, and Zheng]{zhao2021anomaly}
He Zhao, Yuexiang Li, Nanjun He, Kai Ma, Leyuan Fang, Huiqi Li, and Yefeng Zheng.
\newblock Anomaly detection for medical images using self-supervised and translation-consistent features.
\newblock \emph{IEEE Transactions on Medical Imaging}, 40\penalty0 (12):\penalty0 3641--3651, 2021.

\bibitem[Zhou et~al.(2023)Zhou, Pang, Tian, He, and Chen]{zhou2023anomalyclip}
Qihang Zhou, Guansong Pang, Yu Tian, Shibo He, and Jiming Chen.
\newblock Anomalyclip: Object-agnostic prompt learning for zero-shot anomaly detection.
\newblock \emph{arXiv preprint arXiv:2310.18961}, 2023.

\bibitem[Zhu et~al.(2024)Zhu, Cai, Deng, Ooi, and Wu]{zhu2024llms}
Jiaqi Zhu, Shaofeng Cai, Fang Deng, Beng~Chin Ooi, and Junran Wu.
\newblock Do llms understand visual anomalies? uncovering llm's capabilities in zero-shot anomaly detection.
\newblock In \emph{Proceedings of the 32nd ACM International Conference on Multimedia}, pages 48--57, 2024.

\bibitem[Zou et~al.(2022)Zou, Jeong, Pemula, Zhang, and Dabeer]{zou2022spot}
Yang Zou, Jongheon Jeong, Latha Pemula, Dongqing Zhang, and Onkar Dabeer.
\newblock Spot-the-difference self-supervised pre-training for anomaly detection and segmentation.
\newblock In \emph{European Conference on Computer Vision}, pages 392--408. Springer, 2022.

\end{thebibliography}
}

\end{document}